\documentclass[10pt,journal,compsoc]{IEEEtran}

\usepackage{times}
\usepackage{epsfig}
\usepackage{graphicx}
\usepackage{amsmath}
\usepackage{amssymb}
\usepackage{color}

\usepackage{epsfig}
\usepackage{graphicx}
\usepackage{amsmath}
\usepackage{amssymb}
\usepackage{bm}
\usepackage{esvect}
\usepackage{amsbsy}
\usepackage{multirow}
\usepackage[export]{adjustbox}
\usepackage{hhline}
\usepackage{makecell}

\usepackage{xspace}
\usepackage{hyperref}
\makeatletter
\DeclareRobustCommand\onedot{\futurelet\@let@token\@onedot}
\def\@onedot{\ifx\@let@token.\else.\null\fi\xspace}

\def\eg{\emph{e.g}\onedot} 
\def\ie{\emph{i.e}\onedot}

\def\wrt{w.r.t\onedot} 

\makeatother

\newcommand{\bo}[1]{\boldsymbol{#1}}

\ifCLASSOPTIONcompsoc
  \usepackage[nocompress]{cite}
\else
  \usepackage{cite}
\fi
\ifCLASSINFOpdf
\else
\fi

\usepackage{array,booktabs}
\newcolumntype{C}[1]{>{\centering\let\newline\\\arraybackslash\hspace{0pt}}p{#1}}

\begin{document}

%

\title{Forecasting People Trajectories and Head Poses by Jointly Reasoning on Tracklets and Vislets}
%
%
%
%

\author{Irtiza~Hasan,~\IEEEmembership{}
        Francesco~Setti,~\IEEEmembership{}
        Theodore~Tsesmelis,~\IEEEmembership{}
        Vasileios~Belagiannis,~\IEEEmembership{}\\
        Sikandar~Amin,~\IEEEmembership{}
        Alessio~{Del Bue},~\IEEEmembership{}
        Marco~Cristani,~\IEEEmembership{}
        and~Fabio~Galasso~\IEEEmembership{}
\IEEEcompsocitemizethanks{
\IEEEcompsocthanksitem I. Hasan is with Inception Institute of Artificial Intelligence, Abu Dhabi, UAE. This work was done while working at University of Verona, Verona, Italy and Osram GmbH, Munich, Germany.
\protect\\
E-mail: irtiza.hasan@inceptioniai.org
\IEEEcompsocthanksitem F. Setti, and M. Cristani are with the Department of Computer Science, University of Verona, Verona, Italy.

\IEEEcompsocthanksitem T. Tsesmelis and A. Del Bue are with the Italian Institute of Technology, Genova, Italy.
\IEEEcompsocthanksitem T. Tsesmelis, S. Amin, and F. Galasso are with Osram GmbH, Munich, Germany.
\IEEEcompsocthanksitem V. Belagiannis, is with Ulm University, Ulm, Germany. This work was done while working at Osram GmbH.
}
\thanks{\textcolor{red}{ }}}

%
%

\markboth{IEEE Transactions on Pattern Analysis and Machine Intelligence,~Vol.~XX, No.~YY, November~2018}%
{Hasan \MakeLowercase{\textit{et al.}}: Bare Advanced Demo of IEEEtran.cls for IEEE Computer Society Journals}
%



\IEEEtitleabstractindextext{%


\begin{abstract}
\label{sec:abstract}
In this work, we explore the correlation between people trajectories and their head orientations. We argue that people trajectory and head pose forecasting can be modelled as a joint problem. Recent approaches on trajectory forecasting leverage short-term trajectories (\emph{aka} tracklets) of pedestrians to predict their future paths. In addition, sociological cues, such as expected destination or pedestrian interaction, are often combined with  tracklets. In this paper, we propose MiXing-LSTM (MX-LSTM) to capture the interplay between positions and head orientations (vislets) thanks to a joint unconstrained optimization of full covariance matrices during the LSTM backpropagation. We additionally exploit the head orientations as a proxy for the visual attention, when modeling social interactions. MX-LSTM predicts future pedestrians location and head pose, increasing the standard capabilities of the current approaches on long-term trajectory forecasting. Compared to the state-of-the-art, our approach shows better performances on an extensive set of public benchmarks. MX-LSTM is particularly effective when people move slowly, \ie the most challenging scenario for all other models. The proposed approach also allows for accurate predictions on a longer time horizon.


\end{abstract}

}

\maketitle
\newcommand{\todo}[1]{}
\renewcommand{\todo}[1]{{\color{red} TODO: {#1}}}
\newcommand{\myparagraph}[1]{\noindent\textbf{#1}}

\newcommand{\fs}[1]{{\color{blue}[FRANZ: {#1}]}}
\newcommand{\fg}[1]{{\color{blue}[FABIO: {#1}]}}
\newcommand{\figref}[1]{Fig.~\ref{#1}}

\IEEEdisplaynontitleabstractindextext

%
\IEEEpeerreviewmaketitle


%
%
%
%





%
\begin{figure*}[pt!] 
\begin{center}
	\includegraphics[width=1\linewidth]{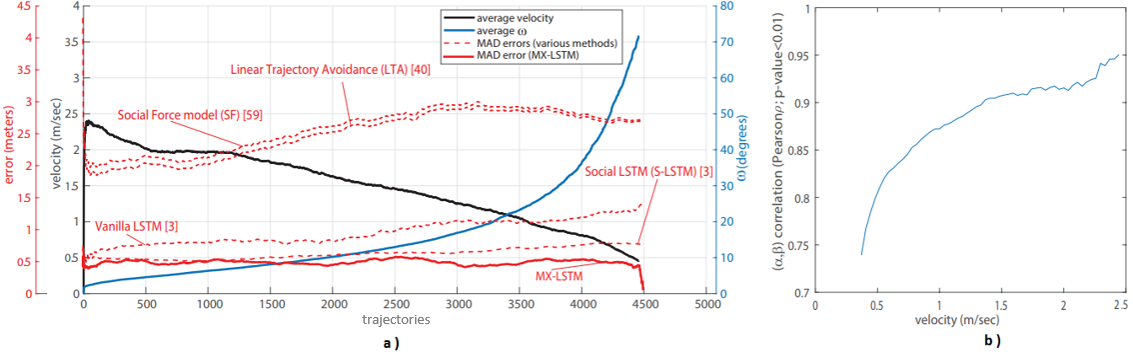}
	\caption{Motivating the MX-LSTM: a) analysis between the angle discrepancy $\omega$ between head pose and movement, the pedestrain velocity and the average errors of different approaches on the UCY sequence~\cite{lerner2007crowds}; b) correlation between movement angle $\beta$ and head orientation angle $\alpha$ when the velocity is varying (better in color).}
	\label{fig:angles}
	\vspace{-10pt}
\end{center}
\end{figure*}
%

\section{Introduction}
\label{sec:intro}

Pedestrian forecasting stands for anticipating the future, based on observations and on prior understanding of the scene and actors. Further to past trajectories, forecasting the position of pedestrians requires therefore an intuition of the people goals~\cite{pellegrini2009iccv}, their social interaction models~\cite{robicquet2016eccv,alahi2016cvpr,SocialGAN18}, the understanding of their behavior~\cite{antonini08behavioralpriors,lee2016predicting,ma2017forecasting} and possible interactions with the scene~\cite{kitani2012activity}. 

Forecasting is important for tracking~\cite{yamaguchi2011cvpr,sadeghian2017tracking,Luo18}, especially in the case of missing or sparse target observations. In addition, it is a crucial compound for early action recognition~\cite{Xing08,Ryoo11,Hoai12} and more in general for surveillance systems~\cite{Collins01,Ferryman00}. Furthermore it is indispensable for deploying autonomous vehicles, which should avoid collisions~\cite{bhattacharyya2017long}, and for conceiving robots, respectful of the human proxemics~\cite{dragan2011manipulation,hall1966hidden,kuderer2012feature,mainprice2016goal,trautman2010iros,ziebart2009planning}.

Forecasting trajectories from images, however, is a complex problem and, probably for this reason, it has only recently emerged as a popular computer vision research topic. In particular, the modern re-visitation of Long Short Term Memory (LSTM) architectures~\cite{hochreiter1997long}, has enabled a leap forward in performance~\cite{ijcai2017-386,su2016crowd,sun20173dof,varshneya2017human,SocialGAN18}. On one side, LSTM has allowed a seamless encoding of the social interplay among pedestrians~\cite{alahi2016cvpr,SocialGAN18}. On the other side, the new systems have abandoned cues demanding \emph{oracle} knowledge, such as the person destination point~\cite{pellegrini2009iccv}, and are therefore causal predictions.

In this paper, we differ from previous approaches, because we additionally leverage the visual attention of people for forecasting, further to their position. We infer their visual attention from their head pose. We are motivated by the strong correlation between the past short-term trajectories of the people (sequences of $(x,y)$ position coordinates, named \emph{tracklets}) and their corresponding sequences of head pan orientations, which we name \emph{vislets}. 
Our novel contribution is supported by several sociological studies~\cite{caminada1980philips,davoudian2012pedestrians,fotios2015using,fotios2015usingII,foulsham2011and,patla2003far,vansteenkiste2013visual}
and here motivated by statistical analysis conducted on the UCY dataset~\cite{lerner2007crowds}, which we report in Section~\ref{sec:motivation}.

This work introduces MiXing LSTM (MX-LSTM), an LSTM-based framework that encodes the relation between the movement of the head and people dynamics. For example, 
it captures the fact that
rotating the head towards a particular direction may anticipate turning and starting to walk
(as in the case of a person leaving a group after a conversation). This is achieved in MX-LSTM by mixing the tracklet and vislet streams in the LSTM hidden state recursion by means of a cross-stream full covariance matrix. During the LSTM  backpropagation, the covariance matrix is constrained to be positive-semidefinite by means of a log-Cholesky parameterization. 
This  model generalizes the approach of~\cite{alahi2016cvpr} (specific to the 2D positions \textit{x,y} of people) to model state variables of dimensions four (position and head pose) and higher.

Vislets allow for a more informative social interplay among people. Instead of considering all pedestrians within a radius, as done in \cite{alahi2016cvpr,varshneya2017human}, here we only consider those individuals whom the person can see. Furthermore MX-LSTM forecasts both tracklets and vislets. Predicting visual attention in crowded scenarios makes a novel frontier for research and new applications.


We have first presented MX-LSTM in \cite{Hasan18}. This paper extends our previous work in four directions:
1) we include a comprehensive evaluation of its performance on the UCY video sequences (Zara01, Zara02 and UCY)~\cite{lerner2007crowds} and on the TownCentre dataset~\cite{benfold2009guiding}, following standard evaluation protocols of trajectory forecasting~\cite{pellegrini2009iccv,alahi2016cvpr,SocialGAN18}. 2) we provide an extensive evaluation with the most recent approaches to show that MX-LSTM retains overall the best performance. MX-LSTM has the ability to forecast people when they are moving slowly, the Achilles heel of all the other approaches proposed so far. 
Additionally, here we provide novel experiments to test its robustness by predicting in the longer-term horizon and by using an estimated (thus noisy) head pose estimator~\cite{lee2015fast}. In particular we quantify the performance of head pose estimates \emph{vs.} manual labels both at training and inference.
3) We verify that vislets help beyond the mere larger model capacity, by testing MX-LSTM with position-related variables replacing vislets. 4) we provide novel qualitative illustrations, detail failure cases; and finally we perform novel simulations, which uncover how the learned head poses affect the people motion.

The rest of the paper is organized as follows: In Section~\ref{sec:motivation}, we motivate the need for MX-LSTM showing the correlation between head pose and trajectories in the most popular forecasting datasets. Section~\ref{sec:prev} presents the related literature. In Section~\ref{sec:our}, we present our MX-LSTM approach. Section~\ref{sec:experiments} illustrates quantitative and qualitative experiments and ablation studies. Finally, Section~\ref{sec:conc} concludes the paper.

\section{Motivation for the MX-LSTM}
\label{sec:motivation}

Intuitively, the head pose of person is a cue for the direction in which she/he moves. However, the literature in trajectory forecasting lacks a quantitative study on the importance of the head pose. Here we examine the common forecasting datasets to study the relationship between the head pose and motion directions. 
In particular, we focus on the UCY dataset~\cite{lerner2007crowds}, composed by the Zara01, Zara02 and UCY sequences, which provides the annotations for the pan angle of the head pose of all the pedestrians. We also consider the Town Center dataset~\cite{benfold2011stable}, where we have manually annotated the head pose, using the same annotation protocol as in~\cite{lerner2007crowds}.

In this section, with specific reference to Figure~\ref{fig:angles}, we present the preliminary analysis and observations, which have motivated the design of our MX-LSTM. We would specifically refer to the UCY video sequence (but similar observations applied to all others).


\vspace{.5em}
\noindent\textbf{1) People watch their steps.}
We show this fact by plotting in Fig.~\ref{fig:angles}a the angular discrepancy $\omega$ (\textit{blue curve}), between the head pose $\alpha$ and the person motion angle $\beta$, against the velocity (\textit{black curve}), intended as the modulus of the motion vector $\vv{\bm{x_{t+1}-x_{t}}}$.

In more details, we have computed the average angular discrepancy $\omega$ for each of the people trajectories of the UCY video sequence (for each trajectory, we average $\omega$ across all frames where it occurs). In Fig.~\ref{fig:angles}a, we have then arranged the trajectories in ascending order (the \textit{x} axis) according to their average discrepancy angle $\omega$ (the \textit{blue} \textit{y}-axis on the sub-figure right side, marked as ``$\omega$''. Please refer to Fig.~\ref{fig:explanations}c, for pictorial illustration of ``$\omega$''). For each trajectory we have then plotted the corresponding average speed (\textit{black curve}), as measured on the black \textit{y}-axis marked as ``velocity''\footnote{We disregard those frames where the average speed of person movement is below 0.45m/sec, since those people do not essentially move and their motion angle $\beta$ can hardly be determined.}.

As it shows from Fig.~\ref{fig:angles}a, 75\% of the people only turn their head by $20^{\circ}$. They watch therefore their steps, especially at higher speeds.


%


\vspace{.5em}
\noindent\textbf{2) Head pose and movements are (statistically) correlated.}
On Fig.~\ref{fig:angles}a, we report the velocity curve (black solid line and axis). To plot this curve, we order all the trajectories with respect to the average speed of each individual.
%
First of all, notice that the $\omega$ and the pedestrian speed are inversely proportional: the alignment between the head pose and the direction of movement is higher when the speed is higher; when the person slows down the head pose is dramatically misaligned.
Secondly, the relation is statistically significant: we consider the Pearson circular correlation coefficient~\cite{jammalamadaka2001topics} between the angles $\alpha_{t}$ and $\beta_t$. Overall, the correlation is 0.83 (p-value$<0.01$), computed for all the frames of the sequences considered for Fig.~\ref{fig:angles}. The plot in Fig.~\ref{fig:angles}b elaborates that the correlation is lower at low velocities, where the discrepancy between the $\alpha_t$ and $\beta_t$ angles is typically higher. 

One of the challenges here, is to investigate whether the dynamic discrepancy between the head pose angle $\alpha_t$ and movement direction $\beta_t$ at different speeds of the human motion can be learned by our proposed MX-LSTM to improve the forecasting. Moreover, MX-LSTM should learn how these relations evolve in time, which has not been investigated yet. In fact, prior work has only addressed single frames.


\vspace{.5em}
\noindent\textbf{3) Forecasting is difficult for pedestrians at low speeds.} 
In Fig.~\ref{fig:angles}a (red lines and red axis), we compare the Mean Average Displacement (MAD) error~\cite{pellegrini2009iccv} of the following approaches: SF~\cite{yamaguchi2011cvpr}, LTA~\cite{trautman2010iros}, vanilla LSTM and Social LSTM~\cite{alahi2016cvpr}, against our proposed MX-LSTM approach (solid red curve).  We notice that lower velocities correspond generally to higher forecasting errors. When people walk slowly, their behavior becomes less predictable, not only due to physical reasons (less inertia), but also behavioral (people walking slowly are usually involved in secondary activities, such as looking around or chatting with others). By contrast, our proposed approach MX-LSTM (solid red curve) performs well even at lower velocities, since it makes use of the evidence from the head pose. MX-LSTM approaches an error close to zero for the nearly static people, as it should ideally be (more details in Sec. \ref{sec:experiments}). 

\vspace{1em}
\noindent Summarizing, the head pose is correlated with the movement. When people move fast, this correlation is stronger and their head pose is largely aligned with the direction of motion. However, when people move slowly, the correlation is weaker (but still significant), and the head pose is drastically misaligned with the movement. This results in higher prediction errors for most state-of-the-art approaches of trajectory forecasting. These facts  justify  and  motivate  our  objective  with  the  MX-LSTM,  to capture the head pose information jointly with the movement and use  it  for  a  better  and  more  uniform trajectory  forecasting,  for people moving at both lower or higher speeds.


%
%

\section{Related work}
\label{sec:prev}


Trajectory forecasting~\cite{becker2018evaluation, morris2008survey} has been traditionally addressed by approaches such as Kalman filter~\cite{kalman1960new}, linear~\cite{mccullagh1989generalized} or Gaussian regression models~\cite{quinonero2005unifying,rasmussen2006gaussian,wang2008gaussian,williams1998prediction}, auto-regressive models~\cite{akaike1969fitting} and time-series analysis~\cite{priestley1981spectral}. 
The main limitation of these approaches is the lack of modelling the human-human interactions~\cite{antonini2006discrete, choi2012unified, choi2014understanding, leal2011everybody, treuille2006continuum}, that instead plays an important role. 
More recent approaches have proposed to use convolutional neural networks~\cite{huang2016deep}, generative models~\cite{gupta2018social} and recurrent neural networks~\cite{alahi2016cvpr} for modelling the trajectory prediction, which also consider the human-human interaction.
We discuss these most recent related approaches in the respective subsections below. 


\noindent\textbf{Human-human interactions.}
Helbing and Molnar~\cite{helbing1995social} have considered for the first time the effect of other pedestrians to the behavior of an individual. The pioneering idea has been further developed by~\cite{lerner2007crowds}, \cite{ma2017forecasting} and \cite{pellegrini2009iccv}, who have respectively introduced a data-driven, continuous and game theoretical model. Notably, these approaches successfully employed the essential cues for track prediction, such as the human-human interaction and people intended destination.
More recent works encode the human-human interactions into a ``social'' descriptor~\cite{alahi2014socially} or propose human attributes~\cite{yi2015understanding} for the forecasting in crowds.
Other related methods~\cite{alahi2016cvpr,varshneya2017human} embed the proxemic reasoning into an LSTM-based predictor. Here the social aspect is implicitly addressed by pooling the hidden LSTM variables of all actors participating in the motion.
Our work mainly differentiates from~\cite{alahi2016cvpr,lerner2007crowds,pellegrini2009iccv,varshneya2017human} because we only consider for interactions those people who are within the cone of attention of the person, (as also verified by psychological studies~\cite{intriligator2001spatial}). 

\noindent\textbf{Destination-focused path forecast.}
%
Path forecasting has also been framed as an inverse optimal control (IOC) problem by Kitani \textit{et.~ al.}~\cite{kitani2012activity}. The follow-up works ~\cite{abbeel2004apprenticeship,ziebart2008maximum} have adopted inverse reinforcement learning and dynamic reward functions~\cite{lee2016predicting} to address the occurring changes in the environment. We describe these approaches as destination-focused, because they require the end-point of the person track to be known. To eliminate this constraint, similar works have relaxed the destination end-point to a set of plausible path ends~\cite{dragan2011manipulation,mainprice2016goal}. By contrast, our approach does not require this information and it is therefore causal (while knowledge of end-point would require knowing the future).


\noindent\textbf{Head pose as social motivation.}
Our interest into the head pose stems from sociological studies such as \cite{caminada1980philips,davoudian2012pedestrians,fotios2015using,fotios2015usingII,foulsham2011and,patla2003far,vansteenkiste2013visual}, whereby the head pose has been shown to correlate to the person destination and pathway. Interestingly, the correlation is higher in the cases of poor visibility, such as at night time, and in general when the person is being busy with a secondary task (\eg bump avoidance) further to the basic walking~\cite{fotios2015using,fotios2015usingII}.
In our experimental studies, we observe that the head pose is correlated with the movement, especially at high velocities, while slowing down this correlation decreases too, but still remains statistically significant. These studies motivate the use of the head pose as proxy to the track forecasting.

There is prior work on estimating the head pose of people in real-time, applicable to people at low resolution~\cite{ba2004probabilistic,gourier2006head,hasan2017tiny,lee2015fast,robertson2006estimating,stiefelhagen1999gaze,tosato2013characterizing}. We leverage these methods within MX-LSTM, to gather the required head pose information from the input frames (just).
To the best of our knowledge, there is no prior work using head pose to forecast the pedestrian trajectories, further to our own. In~\cite{hasan2018seeing}, we integrate the view frustum of attention into an objective energy formulation. By contrast, the proposed LSTM-based framework provides an implicit data-driven joint formulation, which outperforms our previous method. In \cite{Hasan18}, we introduce MX-LSTM for the first time. Here we extend it with novel quantitative and qualitative evidence.

\noindent\textbf{LSTM models.}
LSTM models~\cite{hochreiter1997long} have been employed in tasks where the output is conditioned on a varying number of inputs~\cite{gregor2015draw,vinyals2015show}, notably hand writing generation~\cite{graves2013generating}, tracking~\cite{coskun2017long}, action recognition~\cite{du2015hierarchical, liu2016spatio}, future prediction~\cite{huang2016deep, lee2017desire, srivastava2015unsupervised} and path prediction~\cite{xue2018ss}.

%
%
%

As for trajectory forecasting, Alahi \textit{et.~al.}~\cite{alahi2016cvpr} model the pedestrians as LSTMs that share their hidden states through a ``social'' pooling layer, avoiding to forecast colliding trajectories. This idea has been successfully adopted by~\cite{varshneya2017human}. In~\cite{sadeghian2017tracking}, it has been extended for modeling the tracking dynamics. A similar approach~\cite{ijcai2017-386,su2016crowd} has been embedded directly in the LSTM memory unit as a regularization, which models the local spatio-temporal dependency between neighboring pedestrians. 
In this work, we propose a variant of the social pooling by considering a visibility attention area, defined by the head pose.


In most cases, the training of forecasting LSTMs is driven by the minimization of negative log-likelihoods whereby the probabilities are  Gaussians~\cite{alahi2016cvpr,varshneya2017human} or mixture of Guassians~\cite{graves2013generating}. In general, when it comes to Gaussian 
parameters, only bidimensional data (i.e. $(x,y)$ coordinates) have been considered so far, leading to the estimation of 2 x 2 covariance matrices. These can be optimized without considering the positive semidefinite requirement~\cite{graves2012supervised}, that is one of the most important problems for the covariances obtained by optimization~\cite{pinheiro1996unconstrained} (see Sec.~\ref{sec:lstmopt}). Here, we study the problem of optimizing Gaussian parameters of higher dimensionality for the first time.


\section{Our approach}
\label{sec:our}


In this section, we present \emph{MX-LSTM}. The model may jointly forecast individuals' locations and pose by leveraging the information about the recent history of head positions (\emph{tracklets}) and orientations (\emph{vislets}).
We first  define the concepts of tracklets and vislets (Sec.~\ref{sec:trackvis}); then we describe our proposed formulation of social pooling based on visual frustum of attention (Sec.~\ref{sec:socpool}); finally, we report details about the LSTM formulation (Sec.~\ref{sec:lstmrec}) and model training by optimizing the multidimensional co-variance matrices (Sec.~\ref{sec:lstmopt}).

\subsection{Tracklets and vislets}
\label{sec:trackvis}
\begin{figure}[t]
    \begin{center}
    	\includegraphics[width=1\linewidth]{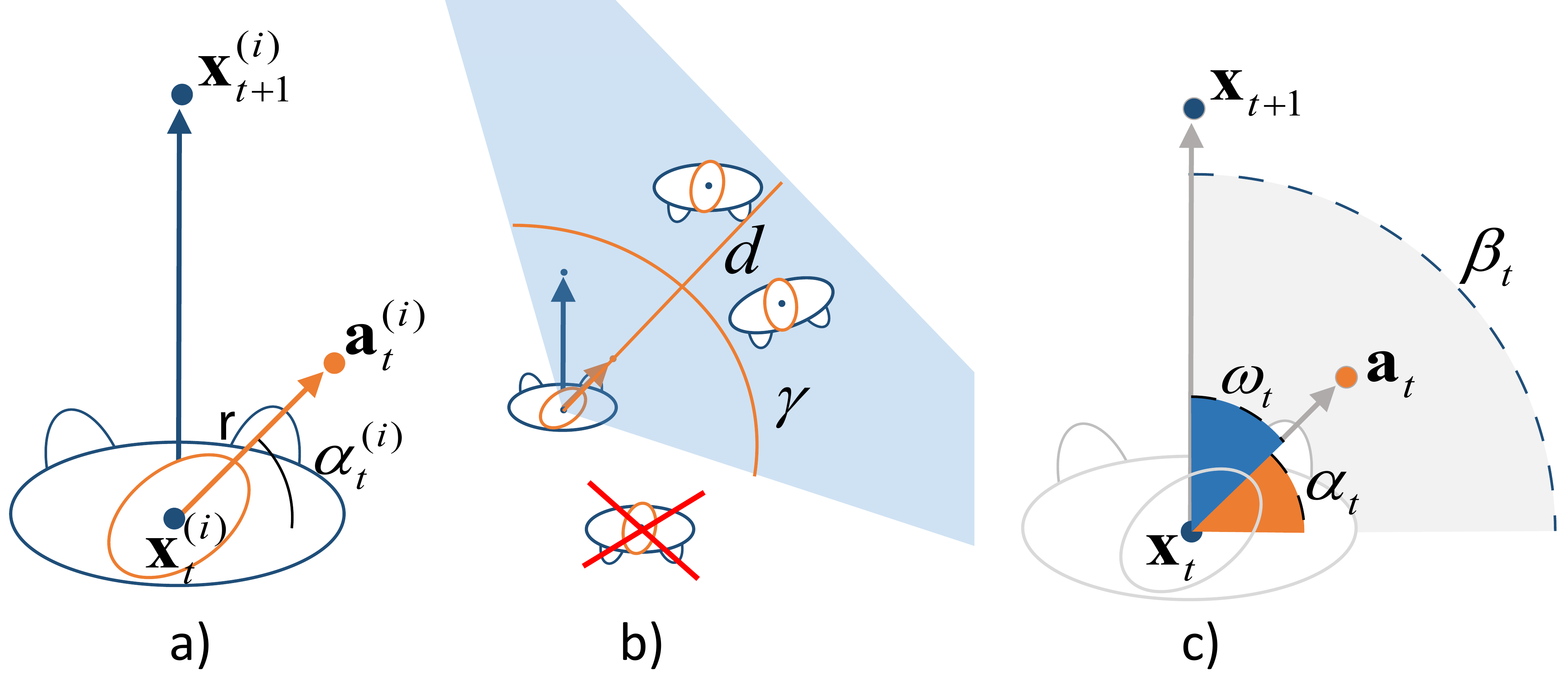}
        \caption{A graphical interpretation of tracklets and viselets. a) tracklets $\mathbf{x}^{(i)}_t$ and $\mathbf{x}^{(i)}_{t+1}$ and vislet anchor point $\mathbf{a}^{(i)}_t$; b) Social pooling leveraging the Visual Frustum of Attention; c) angles for the correlation analysis.}
        \label{fig:explanations}
    \end{center}
\end{figure}

We define as \emph{tracklet} the list of consecutive locations on the ground plane visited by an individual during the last time steps. Formally, the tracklet associated with the $i$-th subject at time $T$ is $\{\mathbf{x}^{(i)}_t\}_{t=1,...,T}$, where $\mathbf{x}^{(i)}_t = (x,y)\in \mathcal{R}^2$.
Similarly, a \emph{vislet} is the list of anchor points located at a fixed distance $r$ from the subject, aligned with its head head orientation. Thus, for subject $i$ at time $T$ the vislet is $\{\mathbf{a}^{(i)}_t\}_{t=1,...,T}$, with $\mathbf{a}^{(i)}_t = (x_t^{(i)}+\cos\alpha_t^{(i)},y_t^{(i)}+\sin\alpha_t^{(i)}) \in \mathcal{R}^2$ (see \figref{fig:explanations}a).

In theory, one could encode the head orientation by means of the pan angle at each time step. We prefer to use anchor points instead, which gives several benefits.
The main advantage of using vislets instead of encoding the head orientation directly with the pan angle, is that this formulation implicitly solve all the issues generated by the discontinuity between 360$^\circ$ and 0$^\circ$.
Moreover, vislets and tracklets have very similar representations, which is very convenient for modeling the interplay of these two components in the MX-LSTM structure.
Please note that the distance $r$ is irrelevant, as long as it is a constant value; in this work we set it at $0.5$m for the sake of visualization.

Our method relies on a set of location and head pose observations to predict tracklets and vislets for the following estimation period. In particular, MX-LSTM mixes together the two streams to understand their relationship, providing a joint prediction.
Accordingly to the trajectory forecasting literature~\cite{alahi2016cvpr,trautman2010iros,yamaguchi2011cvpr}, we consider these observations as provided by an oracle, \emph{i.e.} given by an annotator. To directly compare our approach with the other recent ones, we provide experiments where the past head poses are estimated by a real ``static'' head pose estimator; in this way, MX-LSTM will require no additional effort in annotation with respect to former approaches.


We instantiate an LSTM model for each individual by using two separate embedding functions for tracklets~\eqref{eq:embxy} and vislets~\eqref{eq:embvis}:
\begin{eqnarray}
    \mathbf{e}_{t}^{(x,i)} &=& \phi \left( \mathbf{x}^{(i)}_t,\mathbf{W}_{x} \right) \label{eq:embxy}\\
    \mathbf{e}_{t}^{(a,i)} &=& \phi \left( \mathbf{a}^{(i)}_t,\mathbf{W}_{a} \right) \label{eq:embvis}
\end{eqnarray}
where the embedding function $\phi$ is the linear projection, via the embedding weigths $\mathbf{W}_{(\cdot)}$, into a $D$-dimensional vector, with $D$ the dimension of the hidden space. This is followed by a ReLU activation function.

\subsection{VFoA social pooling}
\label{sec:socpool}
\begin{figure}[t!]
  \centering
  \includegraphics[width=1\linewidth]{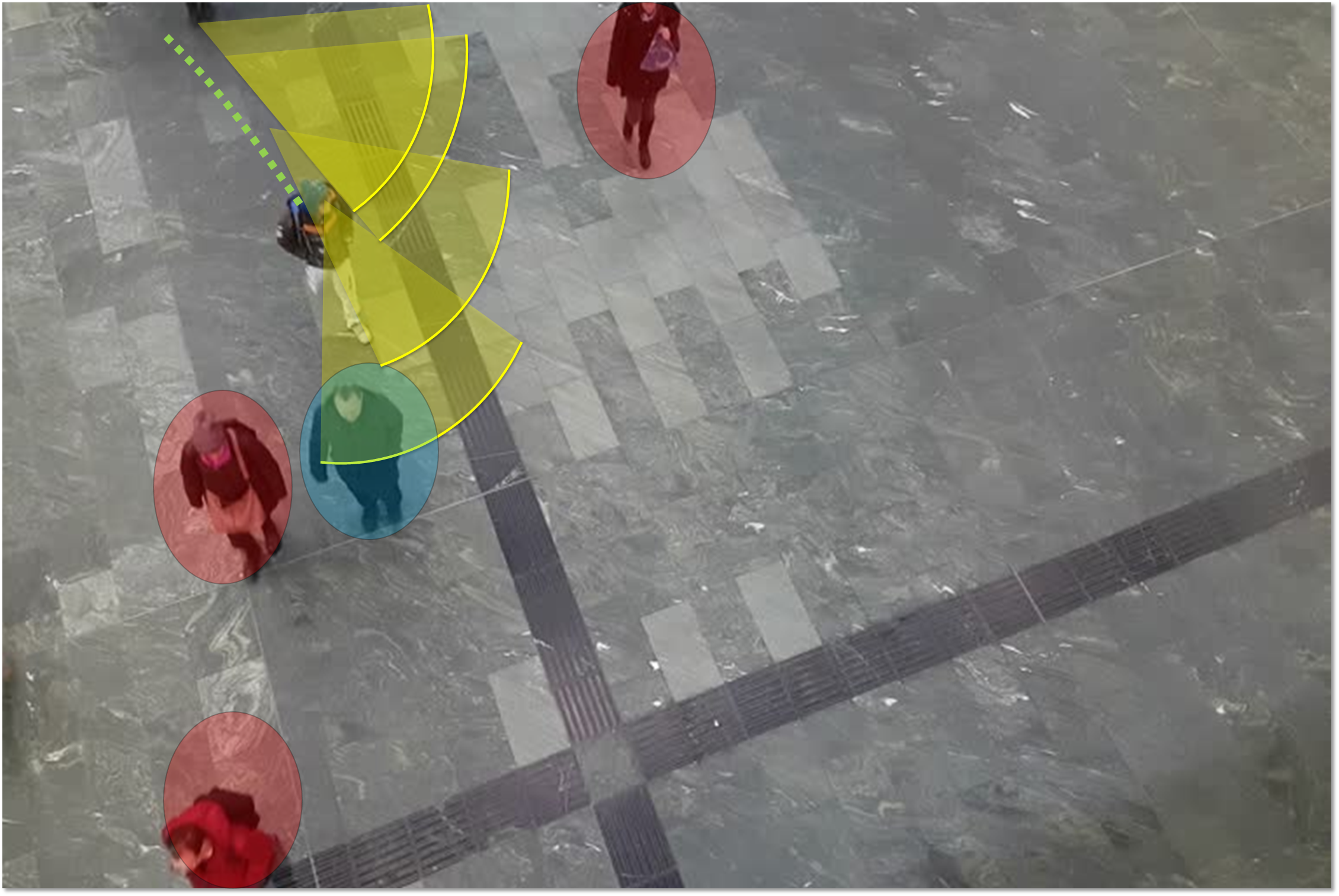}
  \caption{VFOA pooling: For a given subject, he will try to avoid collision with the people who are inside his view frustum (blue circle). Others (red circle), will not influence his trajectory as they are no in his view frustum.}
  \label{fig:vfoaPoolingIllustration}
\end{figure}

The concept of social pooling was first introduced by~\cite{alahi2016cvpr} as an effective way to capture (and embed into an LSTM model) how people move in a crowded space to avoid collisions. In its original form, it is an isotropic area of interest surrounding the target individual. The LSTM hidden variables of the people within the area of interest are pooled, \ie collected to account for the human-human interaction. This formulation implicitly assumes that a person's trajectory is affected not only by the behaviour of people walking in front of him/her, but also by people behind him/her back as also illustrated in \figref{fig:vfoaPoolingIllustration}.
In this paper we upgrade this model by exploiting vislet information, building on the concept of View Frustum of Attention (VFoA), that is a region where the attention of a person is focused, according to its gaze direction. 
We propose to model the VFoA as a circular sector originating in the head position ($\mathbf{x}^{(i)}_t$), aligned with the head pose (\ie towards the anchor point $\mathbf{a}^{(i)}_t$), with a aperture angle $\gamma$; to account for the limitations of human vision in focusing on very far ahead objects, we limit the region with a maximum distance $d$.
We learned both $\gamma$ and $d$ parameters at training time by cross-validation on the training partition of the TownCentre dataset.
A graphical interpretation of the VFoA is provided in \figref{fig:explanations}(b).



Formally, we define an area of interest as the squared region centered at the pedestrian location with size $2d\times 2d$; this area is then divided in a uniform grid of $N_o\times N_o$ cells.
Our VFoA social pooling is a $N_o \times N_o \times D$  tensor $\mathbf{H}$ defined as follows:
%
%
\begin{equation}
    \mathbf{H}^{(i)}_t(m,n,:)=\sum_{j\in \text{VFoA}_i}\mathbf{h}^{(j)}_{t-1},\label{eq:social}
\end{equation}
where the $m$ and $n$ indices run over the $N_o\times N_o$ grid and the condition $j\in \text{VFoA}_i$ is satisfied when the subject $j$ is in the VFoA of subject $i$, $\mathbf{h}$ is the hidden state of the LSTM model.
The pooling vector is then embedded into a $D$-dimensional vector by
\begin{equation}
    \mathbf{e}_{t}^{(H,i)} = \phi({\mathbf{H}^{(i)}_t,\mathbf{W}_H}).\label{eq:pooling}
\end{equation}

\subsection{LSTM recursion}
\label{sec:lstmrec}
The MX-LSTM recursion equation is:
\begin{equation}
    \mathbf{h}^{(i)}_t=LSTM\left(\mathbf{h}^{(i)}_{t-1},\mathbf{e}_{t}^{(x,i)},\mathbf{e}_{t}^{(a,i)},\mathbf{e}_{t}^{(H,i)},\mathbf{W}_{\text{LSTM}}\right).
\end{equation}
The hidden state of the LSTM model projects onto the four dimensional space, representing the Gaussian multi-variate distribution $\mathcal{N}(\mathbf{\mu}^{(i)}_t,\mathbf{\Sigma}^{(i)}_t)$, as follows:
\begin{equation}
    [\mathbf{\mu}^{(i)}_t,\hat{\mathbf{\Sigma}}^{(i)}_t] = \mathbf{W}_o \mathbf{h}^{(i)}_{t-1},
\end{equation}
where $\mathbf{\mu}^{(i)}_t = [\mu^{(x,i)}_t,\mu^{(y,i)}_t,\mu^{(a_x,i)}_t,\mu^{(a_y,i)}_t]$, $\mathbf{\Sigma}^{(i)}_t$ contains the covariances among the $(x,y)$ coordinate distributions of the tracklets and the vislets, and $\hat{\mathbf{\Sigma}}^{(i)}_t$ is its vectorized version.
The distribution is then sampled to generate the joint prediction of tracklets and vislet points $[\mathbf{\hat x}_t,\mathbf{\hat a}_t]$, allowing us to simultaneously forecast trajectries and head poses.

At training time, we estimate the weights of the LSTM by minimizing the multivariate Gaussian log-likelihood for the each trajectory. The loss function is
\begin{eqnarray}
  L^i(\mathbf{W}_{x},\mathbf{W}_{a},\mathbf{W}_H,\mathbf{W}_{\text{LSTM}},\mathbf{W}_{o}) &=&\nonumber\\  -\sum_{T_{obs}+1}^{T_{pred}}log\left( P([\mathbf{x}^{(i)}_t, \mathbf{a}^{(i)}_t],\mathbf{\mu}^{(i)}_t,\mathbf{\Sigma}^{(i)}_t)\right),
  \label{eq:LLog}
\end{eqnarray}
where $T_{obs}$ is the last frame of the observation period, while $T_{obs}+1,\ldots,T_{pred}$ are the time frames for which we provide a prediction. The loss of Eq.~\eqref{eq:LLog} is minimized over all the training sequences. To prevent overfitting, we additionally include an $l_2$ regularization term.

\subsection{MX-LSTM optimization}
\label{sec:lstmopt}
As shown in Eq.~\eqref{eq:LLog}, the optimization procedure provides the weight matrices of the MX-LSTM, which in turn produces the set of Gaussian parameters, including the full covariance $\mathbf{\Sigma}$. The latter is needed to enforce the LSTM in encoding the relations among the $(x,y)$ coordinate distributions of tracklets and vislets, which we already discussed in Sec.~\ref{sec:motivation}.
In principle, one may have simply captured the correlation between the walking direction and head pose in order to model drifts in the trajectory, but we are interested in letting the MX-LSTM analyze also how the head pose (pan angle) influences the length of the spatial step, that is the velocity. In other words, we want the MX-LSTM to be able to capture whether a particular head pose dynamics could accelerate or slow down the motion, thus letting the machine forecast the joint behavior.

The estimation of a full covariance matrix as the result of an optimization procedure over a generic objective function, like the log-likelihood of \eqref{eq:LLog}, is a difficult numerical problem~\cite{pinheiro1996unconstrained}.
The main reason is that one must guarantee that the resulting estimate is a proper covariance matrix, \ie a positive semi-definite (p.s.d.) matrix.
For this reason, LSTMs with log-likelihood loss functions over Gaussian distributions have been restricted so far to two dimensions, using a simple Gaussian~\cite{alahi2016cvpr}, or mixture of Gaussian distributions. The $2\times2$ covariance matrices have been obtained by optimizing the scalar correlation index $\rho_{x,y}$, which becomes the covariance term of $\mathbf{\Sigma}$ with $\sigma_{x,y}=\rho_{x,y}\sigma_x\sigma_y$ ~\cite{graves2013generating}.

In case of higher dimensional problems, pairwise correlation terms cannot be optimized for building $\mathbf{\Sigma}$, since the optimization process for each correlation term is independent from each other.
At the same time, the positive-definiteness is a simultaneous constraint on multiple variables~\cite{pourahmadi2011covariance}.
In practice, if we consider three variables $x$, $y$ and $z$, learning $\rho_{x,y}$ and $\rho_{x,z}$ are two independent procedures, despite that they act on the common distribution over $x$.
This lacks of coordination generates matrices far from being p.s.d. and thus requiring a further correction procedure, It usually consists of projecting the estimated matrix into the closest p.s.d. matrix based on a cost function of the Frobenious norm~\cite{boyd2005least,higham1988computing}.
This procedure is very expensive~\cite{pinheiro1996unconstrained}, and difficult to be embedded into the LSTM optimization process~\cite{dennis1996numerical}, where nonlinearities due to the embedding weights make the analytical derivation hard to formulate. So far, there is not any LSTM loss that involved full covariances of dimension higher than $2$.

Our solution involves unconstrained optimization; we use an appropriate Cholesky parameterization of the matrix to be learned that enforces the positive semi-definite constraint, dramatically improving the convergence properties of the optimization algorithm~\cite{pourahmadi2011covariance}.
We refer the interested reader to \cite{cherian2012jensen} for more details on how to cope with distance measures on covariance matrices.
Let us consider $\mathbf{\Sigma}$ a semi-definite positive $n\times n$ (in our case, $n=4$) covariance matrix. Since $\mathbf{\Sigma}$ is symmetric by definition, only $n(n+1)/2$ parameters are required to represent it. The Choleski factorization is given by:
\begin{equation}
    \mathbf{\Sigma}=\mathbf{L}^T\mathbf{L},\label{Eq:Chole}
\end{equation}
where $\mathbf{L}$ is a $n\times n$ upper triangular matrix. The optimization process focuses on finding the  $n(n+1)/2$ distinct scalar values for $\mathbf{L}$, which we then solve for the covariance, as for Eq.~\eqref{Eq:Chole}.
The main problem with the Cholesky factorization is non-uniqueness: any matrix obtained by multiplying a subset of the rows of $\mathbf{L}$ by -1 is still a valid solution. As a consequence, non-uniqueness makes the problem ill-posed and hinders optimization convergence.
The simplest way to enforce the matrix $\mathbf{L}$ to be unique is to add the constraint that all the diagonal elements must be positive.
To this end, the Log-Cholesky parameterization~\cite{pourahmadi2011covariance} assumes that the values found by the optimizer of the main covariance diagonal are the log of the values of $\mathbf{L}$.
Formally, the values found by the optimizer can be written as:
\begin{equation}
\mathbf{\theta}_{L}= \begin{bmatrix}
    \log l_{1,1}       & l_{1,2} & l_{1,3} & l_{1,4} \\
     0      & \log l_{2,2} & l_{2,3} &   l_{2,4} \\
    0& 0& \log l_{3,3}  & l_{3,4}\\
    0& 0&0 & \log l_{4,4}
\end{bmatrix}.
\end{equation}
In practice, after the estimation of  $\mathbf{W}_{x}$, $\mathbf{W}_{a}$, $\mathbf{W}_H$, $\mathbf{W}_{\text{LSTM}}$, $\mathbf{W}_{o}$ parameters, the values of $\mathbf{\theta}_{L}$ are extracted by
\begin{equation}
    [\mathbf{\mu}^{(i)}_t,\hat{\mathbf{\theta}_{L}}^{(i)}_t] = \mathbf{W}_o \mathbf{h}^{(i)}_{t-1}, \label{eq:params}
\end{equation}
where $\hat{\mathbf{\theta}_{L}}$ is the vectorized version of $\mathbf{\theta}_{L}$. Then, the diagonal values of $\mathbf{\theta}_{L}$ are exponentiated to form $\mathbf{L}$ and obtaining $\mathbf{\Sigma}$ through Eq.~\eqref{Eq:Chole}.

\section{Experiments}
\label{sec:experiments}

\begin{table*}[t]
  \centering
  \caption{Mean and Final Average Displacement errors (in meters) for all the methods on all the datasets. The first 6 columns are the comparative methods and our proposed model trained and tested with GT annotations. MX-LSTM-HPE  is our model 
tested with the output of a real head pose estimator~\cite{hasan2017tiny}. The last 3 columns are variations of our approach trained and tested on GT annotations. 
%
%
}
  \label{tab:res-oracle}
  \resizebox{.999\textwidth}{!}{
  \begin{tabular}{|c|l *{10}{|C{.075\textwidth}}|}
    \hline
    Metric & Dataset    &  SF~\cite{yamaguchi2011cvpr}  &  LTA~\cite{pellegrini2009iccv}   &  Vanilla LSTM~\cite{alahi2016cvpr}  & Social LSTM~\cite{alahi2016cvpr} & Social GAN~\cite{SocialGAN18} & MX-LSTM  & {\it MX-LSTM-HPE } & Individual MX-LSTM  &  NoFrustum MX-LSTM  &  BD- MX-LSTM   \\
    \hline \hline
    \multirow{5}{*}{\centering MAD} & Zara01 & 2.88 & 2.74 & 0.90 & 0.68 & \textit{0.48} & \textbf{0.59} & {\emph{0.66}} & 0.63 & 0.63 & 0.60 \\
    \cline{2-12}
    & Zara02     & 2.32 & 2.23 & 1.09 & 0.63 & \textit{0.44} & {\bf 0.35} & {\emph{0.37}} & 0.72 & 0.36 & 0.41 \\
    \cline{2-12}
    & UCY        & 2.57 & 2.49 & 0.67 & 0.62 & \textit{0.65} & {\bf 0.49} & {\emph{0.55}} & 0.53 & 0.51 & 0.54 \\
    \cline{2-12}
    & TownCenter &9.35 &9.14   & 4.62 & 1.96 & \textit{1.60} & {\bf 1.15} & {\emph{1.21}} & 2.09 & 1.70 & 1.40 \\
    \hline \hline
    \multirow{5}{*}{\centering FAD} & Zara01 & 5.55 & 5.55 & 1.85 & 1.53 & \textit{1.04} & \textbf{1.31} & {\emph{1.43}} & 1.37 & 1.40 & 1.51 \\
    \cline{2-12}
    & Zara02     & 4.35 &  4.35 & 2.15 & 1.43 & \textit{0.95} & {\bf 0.79} & {\emph{0.82}} & 1.56 & 0.84 & 1.00 \\
    \cline{2-12}
    & UCY        & 4.62 &  4.66 & 1.39 & 1.40 & \textit{1.36} & {\bf 1.12} & {\emph{1.20}} & 1.16 & 1.15 & 1.23 \\
    \cline{2-12}
    & TownCenter & 16.01 & 16.08 & 8.26 & 3.96 & \textit{3.50} & {\bf 2.30} & {\emph{2.38}} & 4.00 & 3.40 & 2.90 \\
    \hline
  \end{tabular}
    }
\end{table*}


To validate the proposed approach we perform both qualitative and quantitative evaluations.
%
We report experiments on two public datasets, namely \emph{UCY}~\cite{lerner2007crowds} and \emph{TownCentre}~\cite{benfold2011stable} datasets.
We compare our model with one baseline, \ie a standard LSTM model that only accounts for pedestrian positions (Vanilla LSTM), and four state-of-the-art approaches: Social Force model (SF)~\cite{yamaguchi2011cvpr}, Linear Trajectory Avoidance (LTA)~\cite{pellegrini2009iccv}, Social LSTM (S-LSTM)~\cite{alahi2016cvpr} and Social GAN~\cite{SocialGAN18}.
Here we also investigate three variations of the MX-LSTM model to capture the net contributions of the different parts that characterize our approach.
Moreover, we investigate the effect of changing the observation period and the forecasting horizon, illustrating how head pose plays a pivotal role for the long term forecasting.
Lastly, we 
analyze whether one can substitute the ground-truth head pose information with more accessible proxies, such as the pace direction or head pose estimates, as provided by a detector.
On a qualitative evaluation, we show the interplay between tracklets and vislets that the MX-LSTM has learnt.

\subsection{Implementation details}
\label{sec:impl}

We implemented the MX-LSTM model and all models of the ablation study in Tensorflow. All models have been trained with learning rate of 0.005 along with the RMS-prop optimizer. We set the embedding dimension for spatial coordinates and vislets to 64 and the hidden state dimension is D = 128. We compute the social pooling on a grid of $32\times 32$ cells \eqref{eq:social}, corresponding to 4 meters.
The view frustum aperture angle has been cross-validated on the training partition of the TownCentre and kept fixed for the remaining trials ($\gamma=40^{\circ}$), while the depth $d$ is simply bounded by the social pooling grid. Training and testing has been accomplished with a GPU NVIDIA GTX-1080 for all evaluations.

\subsection{Evaluation Protocol}
\label{sec:evaluation}

We report experiments on two public datasets, namely \emph{UCY}~\cite{lerner2007crowds} and \emph{TownCentre}~\cite{benfold2011stable} datasets. The UCY dataset is composed of three sequences (\emph{Zara01}, \emph{Zara02}, and \emph{UCY}), taken in public spaces from top-view. In Table~\ref{tab:stat}, the statistics for each dataset are compared. For all sequences, the manual annotation of the people position and head pose are available (we have annotated the head pose for the TownCenter and made them available at 
\url{https://github.com/hasanirtiza/MX-LSTM/blob/master/data/} ).

\begin{table}[t!]
  \centering
  \caption{Dataset Statistics}
  \label{tab:stat}
  \resizebox{\linewidth}{!}{
  \begin{tabular}{|l *{4}{|C{.075\textwidth}}|}
    \hline
    Dataset    & Number of  frames & Number of pedestrians & Pedestrians per frame & Average trajectories \\ \hline
    Zara01     & 8,670          & 148         & 6              & 339       \\ \hline
    Zara02     & 10.513         & 204         & 9              & 467       \\ \hline
    UCY        & 5,405          & 434         & 32             & 404       \\ \hline
    TownCentre & 4,500          & 230         & 16             & 310       \\ \hline
  \end{tabular}
  }
\end{table}

The evaluation protocol follows the standard procedure for trajectory forecasting that is used in the literature~\cite{pellegrini2009iccv,alahi2016cvpr}. We first downsample the videos at 2.5fps, then we observe tracklets and vislets for 8 frames, and we predict both locations and head poses for the following 12 time steps.  The observation period is $3.2s$ and the forecasting horizon is $4.8s$.
Experiments with different time horizons are reported in the ablation study (Sec.~\ref{sec:ablation}). According to the standard protocol, we use annotations during the observation period. Since we use additional information with respect to most of the related approaches (\ie head poses), we perform an evaluation with the output of a real head pose estimator as well (Sec.~\ref{sec:ablation}).

For the three UCY sequences we train three models, where we use two sequences for training and the remaining for testing. For the TownCentre dataset, the model has been trained and tested on the provided data splits.

Regarding the evaluation metrics of the trajectory forecasting, we consider the \emph{Mean Average Displacement} (MAD) error, \ie the average Euclidean distance between all the predicted and ground-truth pedestrian locations. The \emph{Final Average Displacement} (FAD) error, \ie the Euclidean distance between the last predicted location of each trajectory and the corresponding manually annotated point, is employed as well. 
Lastly, we evaluate the performance of the head pose predictions in terms of mean angular error $e_{\alpha}$, which is the mean absolute difference between the estimated pose and the annotated ground truth.

\subsection{Comparison with Prior Art}
\label{sec:sota}

We compare our model against a baseline Vanilla LSTM model, which only uses pedestrian positions, and four state-of-the-art approaches: Social Force model (SF)~\cite{yamaguchi2011cvpr}, Linear Trajectory Avoidance (LTA)~\cite{pellegrini2009iccv}, Social LSTM (S-LSTM)~\cite{alahi2016cvpr} and Social GAN~\cite{SocialGAN18}.

Note that the Social GAN~\cite{SocialGAN18} uses ground-truth trajectories during the prediction interval. At test time, the Social GAN~\cite{SocialGAN18} model predicts 20 trajectories and uses the $L_2$ distance \wrt the ground-truth trajectory to select the best one. Although this protocol makes the comparison with all other approach unfair, we include it in the results for the sake of completeness.


Comparative results are reported in Table~\ref{tab:res-oracle}.
The MX-LSTM outperforms the state-of-the-art methods across all sequences on both metrics, except for Zara01, where it underperforms the Social GAN. Overall, MX-LSTM achieves an average improvement of 23.3\% over the second best performer Social GAN. The highest relative gain is achieved in the UCY sequence and TownCentre dataset, where we achieve a MAD error of 0.49 and 1.15 respectively, improving on Social GAN by ~24\% and ~28\% respectively. We explain the larger relative improvement by the increased difficulty of the complex non-linear people paths, in which case the visual attention turns out an important cue. In UCY and TownCenter, people stand in conversational groups, others walk by closely, while some of them slow down to look at the shop windows. We provide quantitative examples of these complex motions in \figref{fig:angles}.

Note that some of the evaluated methods require additional input data: both SF and LTA require the destination point of each individual, while SF additionally requires the social group annotations.
Ours uses the manually labelled (ground-truth) head poses, which are provided to the algorithm (only) in the observation period (before the forecast). We discuss in the next subsection whether this manual annotation is really needed.

\subsubsection{Effect of head pose estimator}
\label{sec:exp-hpe}

Here we analyze the effect on performance, at inference time, of adopting a head pose estimation algorithm~\cite{hasan2017tiny} during the observation period (prior to forecasting), instead of the ground-truth head poses.

We automatically estimate the head bounding box given the feet positions on the floor plane, assuming an average person being $1.80$m tall. Then, we apply the head pose estimator of~\cite{hasan2017tiny} that provides continuous angles for the pan orientation. At inference time, this data is used as input to this variant, which we name ``MX-LSTM-HPE''. 

Results in Table~\ref{tab:res-oracle} illustrate that the performance of MX-LSTM-HPE is in average 9\% worse than MX-LSTM. The importance of the head pose estimate quality for forecasting is therefore notable, which makes future research on head pose an indispensable requirement. Note from Table~\ref{tab:res-oracle} that the results of MX-LSTM-HPE are still better than other techniques across all sequences, with the exception of Social GAN~\cite{SocialGAN18}, outperforming our approach on the UCY sequence.


\subsection{Ablation Study}
\label{sec:ablation}

We analyse the net contribution of different parts of the proposed approach by investigating three variations of our model: namely \emph{Block-Diagonal}, \emph{NoFrustum} and \emph{Individual} MX-LSTM.

\vspace{.5em}
\noindent\textbf{Block-Diagonal MX-LSTM (BD-MX-LSTM)}: This studies the importance of estimating full covariances to understand the interplay between tracklets and vislets, rather than modelling each of them as a separate probability distribution.
%
Essentially, instead of learning the $4\times4$ full covariance matrix $\mathbf{\Sigma}$, BD-MX-LSTM estimates two separate bidimensional covariances $\mathbf{\Sigma}_x$ and $\mathbf{\Sigma}_a$ for the trajectory and the vislet modeling, thus neglecting the cross-stream covariance. Each $2\times2$ covariance is estimated employing two variances $\sigma_1, \sigma_2$ and a correlation terms $\rho$ as presented in~\cite{graves2013generating}. 
%
The equations that differ from the proposed MX-LSTM are Eq.~\eqref{eq:LLog} and Eq.~\eqref{eq:params}, which become:
\begin{eqnarray}
L^i(\mathbf{W}_{x},\mathbf{W}_{a},\mathbf{W}_H,\mathbf{W}_{\text{LSTM}},\mathbf{W}_{o}) &=&\nonumber\\  -\sum_{T_{obs}+1}^{T_{pred}}log\left( P([\mathbf{x}^{(i)}_t]^T|\bo{\mu}^{(x,i)}_t,\bo{\sigma}^{(x,i)}_t,\rho^{(x,i)}_t \right)+\!\!\\\nonumber
log\left(P([\mathbf{a}^{(i)}_t]^T|\bo{\mu}^{(a,i)}_t,\bo{\sigma}^{(a,i)}_t,\rho^{(a,i)}_t\right),
\end{eqnarray}
where $\bo{\mu}^{(x,i)}_t=[{}_x\mu^{(x,i)}_t,{}_y\mu^{(x,i)}_t]$ and the same apply for the variance vector;
\begin{equation}
[\bo{\mu}^{(x,i)}_t,\bo{\sigma}^{(x,i)}_t,\rho^{(x,i)}_t,\bo{\mu}^{(a,i)}_t,\bo{\sigma}^{(a,i)}_t,\rho^{(a,i)}_t]^T \!=\! \mathbf{W}_o \mathbf{h}^{(i)}_{t-1}\label{eq:params2}.
\end{equation}

\vspace{.5em}
\noindent\textbf{NoFrustum MX-LSTM}:Tthis variant reduces MX-LSTM to the social pooling of~\cite{alahi2016cvpr}, i.e.\ pooling for hidden states $\{\mathbf{h}^{j}_t\}$ from the entire area around each individual. NoFrustum MX-LSTM neglects the visual frustum of attention and does not select the people to pool from based on it. Also people behind the person would therefore influence the next step forecasting.

\vspace{.5em}
\noindent\textbf{Individual MX-LSTM}: In this case, no social pooling is taken into account. In more detail, the embedding operation of Eq.~\eqref{eq:pooling} is removed, and the weight matrix $\mathbf{W}_H$ vanishes. In practice, this variant learns independent models for each person, each one considering the tracklet and vislet points.

\vspace{1em}
The last three columns of Table~\ref{tab:res-oracle} report numerical results for the three MX-LSTM variants. The main facts that emerge are: 1) the highest variations are with the Zara02 sequence, where MX-LSTM doubles the performances of the worst approach (Individual MX-LSTM); 2) the worst performing is in general Individual MX-LSTM, showing that social reasoning is indeed needed; 3) social reasoning is systematically improved with the help of the vislet-based view-frustum; 4) full covariance estimation has a role in pushing down the error which is already small with the adoption of vislets.


Summarizing the results so far, having vislets as input allows to definitely increase the trajectory forecasting performance. Vislets should be used to understand social interactions with social pooling, by building a view frustum that tells which are the people currently observed by each individual. All of these features are effectively and efficiently implemented within MX-LSTM. Note in fact that the training time is not affected by whether social pooling is included or not.

Again, although the complete method always outperforms all the competitors, the highest improvement is on the TownCentre sequence. In our opinion this is due to the different level of complexity in the data, indeed most of the trajectories in UCY sequences are relatively linear, with poor social interactions, while in TownCentre there are many interactions, such as forming and splitting groups and crossing trajectories.
For the same reason, this is the dataset where the introduction of the view frustum in the pooling of social interactions gives the highest benefits. By contrast, in all other sequences but Zara01, decoupling the covariance matrix into a block diagonal matrix neglecting the interplay of position and gaze (BD-MX-LSTM) leads to a sensitive increase in the prediction error; this proves the tight relation between the head orientation and the motion of an individual.

\subsection{Head Pose Forecasting}
Our MX-LSTM model also provides a forecast of the head pose of each individual at each frame, for the first time. We evaluate the performances of this estimation in terms of mean angular error $e_{\alpha}$, i.e.\ the mean absolute difference between the estimated pose (angle $\alpha_{t,\cdot}$ in \figref{fig:explanations}c) and the annotated ground truth.
$e_{\alpha}$ expresses how much the direction in which an individual is looking at a particular time instant is different from the true one. This error measure is independent from the error in the predicted position. In other words, $e_{\alpha}$ measures the error in the gaze forecasting.

\begin{table}[t!]
  \centering
  \caption{Mean angular error (in degrees) for the state-of-the-art head pose estimator~\cite{hasan2017tiny}, and our model fed with manual annotations (MX-LSTM) and estimated values (MX-LSTM-HPE).}
  \label{tab:res-angle}
  \resizebox{\linewidth}{!}{
  \begin{tabular}{|l *{4}{|C{.11\textwidth}}|}
    \hline
    Metric         & HPE~\cite{hasan2017tiny} & MX-LSTM & MX-LSTM-HPE \\
    \hline
    Zara01         & 14.29 & 12.98 & 17.69 \\
    \hline
    Zara02         & 20.02 & 20.55 & 21.92 \\
    \hline
    UCY            & 19.90 & 21.36 & 24.37 \\
    \hline
    TownCentre     & 25.08 & 26.48 & 28.55 \\
    \hline
  \end{tabular}
  }
\end{table}

Table~\ref{tab:res-angle} reports numerical results of the static head pose estimator~\cite{lee2015fast} (HPE), the proposed model fed with manually annotated head poses (MX-LSTM) and with the output of HPE (MX-LSTM-HPE) during the observation period.
In all the cases our forecast output is comparable with the one of HPE, but in our case we do not use appearance cues -- i.e. we do not look at the images at all.
In the case of Zara01, the MX-LSTM is even better that the static prediction, which highlights the forecasting power of our model. In our opinion, this is due to the fact that in this sequence trajectories are mostly linear and that people are walking fast, with their heads mostly aligned with the direction of motion. When providing the MX-LSTM model with the estimations during the observation period, the angular error increases, as expected, but the error remains limited.

\subsection{Time Horizon Effect}
\label{sec:horizon}


To investigate how MX-LSTM performs for longer time horizons we conduct an experimental evaluation where we increment the prediction interval from 12 (standard evaluation protocol) to 32 frames with a step size of 4, keeping the observation interval fixed at 8 frames. We evaluated approaches on UCY, Zara01 and Zara02, since most trajectories on TownCenter last less than 24 frames.
We use MAD to report the error.
As shown in Table~\ref{table:table1Ab}, MX-LSTM is well capable of handling longer time horizons. MX-LSTM outperforms all other approaches on all prediction interval, which demonstrates its robustness.
Based on these results, we argue that reasoning on the head pose becomes even more important when forecasting in the longer term. Overall, the ranking is preserved and MX-LSTM remains the best performer. Additionally, in Table \ref{table:tableAdded} we also evaluated approaches that are not causal and require ground-truth information during inference time as well. It can be seen that although Social GAN \cite{SocialGAN18} relies on ground-truth information to select the best track during inference, it outperforms MX-LSTM only on Zara01 dataset. 

\begin{table}[t]
  \centering
  \caption{Mean Average Displacement (MAD) error when changing the forecasting horizon. Observation interval is kept constant at 8 frames.}
  \label{table:table1Ab}
  \resizebox{\linewidth}{!}{
    \begin{tabular}{|c *{7}{|C{.073\textwidth}}|}
    \hline
    Dataset & Forecasting horizon & Vanilla LSTM & Social LSTM & MX-LSTM & Individual MX-LSTM\\
    \hline \hline
    \multirow{4}{*}{\vspace{-2.5em}\textbf{Zara 01}} &
    \textbf{\underline{H = 12}} & 0.90 & 0.68 & \textbf{0.59} & 0.72\\
    \cline{2-6}
    & \textbf{H = 16} & 1.21 & 1.00 & \textbf{0.87} & 1.05\\
    \cline{2-6}
    & \textbf{H = 20} & 1.70 & 1.43 & \textbf{1.21} & 1.44\\
    \cline{2-6}
    & \textbf{H = 24} & 2.30 & 1.94 & \textbf{1.55} & 1.85 \\
    \cline{2-6}
    & \textbf{H = 28} & 3.07 & 2.35 & \textbf{1.92} & 2.47\\
    \cline{2-6}
    & \textbf{H = 32} & 4.11 & 2.85 & \textbf{2.40} & 3.14\\
    \hline \hline
    \multirow{4}{*}{\vspace{-2.5em}\textbf{Zara 02}} &
    \textbf{\underline{H = 12}} & 1.09 & 0.63 & {\bf 0.35} & 0.63\\
    \cline{2-6}
    & \textbf{H = 16} & 1.62 & 0.90 & {\bf 0.53} & 1.09\\
    \cline{2-6}
    & \textbf{H = 20} & 2.19 & 1.24 & {\bf 0.71} & 1.43\\
    \cline{2-6}
    & \textbf{H = 24} & 2.75 & 1.59 & {\bf 0.90} & 1.83\\
    \cline{2-6}
    & \textbf{H = 28} & 3.31 & 2.00 & {\bf 1.16} & 2.25\\
    \cline{2-6}
    & \textbf{H = 32} & 3.86 & 2.41 & {\bf 1.40} & 2.67\\
    \hline \hline
    \multirow{4}{*}{\vspace{-2.5em}\textbf{UCY}} &
    \textbf{\underline{H = 12}} & 0.67 & 0.62 & {\bf 0.49} & 0.53\\
    \cline{2-6}
    & \textbf{H = 16} & 0.90 & 0.90 & {\bf 0.70} & 0.77\\
    \cline{2-6}
    & \textbf{H = 20} & 1.19 & 1.08 & {\bf 0.95} & 1.01\\
    \cline{2-6}
    & \textbf{H = 24} & 1.52 & 1.36 & {\bf 1.22} & 1.27\\
    \cline{2-6}
    & \textbf{H = 28} & 1.87 & 1.66 & {\bf 1.50} & 1.53\\
    \cline{2-6}
    & \textbf{H = 32} & 2.24 & 1.99 & {\bf 1.80} & 1.83\\
    \hline
    \end{tabular}
  }
\end{table}

\begin{table}[t]
  \centering
  \caption{Comparison of MX-LSTM against techniques which leverage ground-truth information from future frames.  Mean Average Displacement (MAD) error when changing the forecasting horizon. Observation interval is kept constant at 8 frames. Note that this comparison is unfair to MX-LSTM, which only uses information from past frames.}
  \label{table:tableAdded}
  \resizebox{0.9\linewidth}{!}{
    \begin{tabular}{|c *{6}{|C{.073\textwidth}}|}
    \hline
    Dataset & Forecasting horizon & Social GAN & LTA & SF & MX-LSTM\\
    \hline \hline
    \multirow{4}{*}{\vspace{-2.5em}\textbf{Zara 01}} &
    \textbf{\underline{H = 12}} &\textbf{ 0.48} & 2.74 & 2.88 & 0.59\\
    \cline{2-6}
    & \textbf{H = 16} & \textbf{0.68} & 3.60& 3.65 & 0.87\\
    \cline{2-6}
    & \textbf{H = 20} & \textbf{0.94} &  4.20 & 4.21 & 1.21\\
    \cline{2-6}
    & \textbf{H = 24} & \textbf{1.26} & 4.60& 4.61 & 1.55 \\
    \cline{2-6}
    & \textbf{H = 28} & \textbf{1.66} &  4.70 & 4.74 & 1.92\\
    \cline{2-6}
    & \textbf{H = 32} & \textbf{2.20} &  4.74 &4.82 & 2.40 \\
    \hline \hline
    \multirow{4}{*}{\vspace{-2.5em}\textbf{Zara 02}} &
    \textbf{\underline{H = 12}} & 0.44 & 2.23 & 2.32 & {\bf 0.35} \\
    \cline{2-6}
    & \textbf{H = 16} & 0.60 & 3.70 & 3.80 & {\bf 0.53} \\
    \cline{2-6}
    & \textbf{H = 20} & 0.76 & 4.15 & 4.20 & \textbf{0.71}\\
    \cline{2-6}
    & \textbf{H = 24} & 0.95 & 4.30 & 4.37 & \textbf{0.90}\\
    \cline{2-6}
    & \textbf{H = 28} & 1.17 & 4.58 & 4.66 & \textbf{1.16}\\
    \cline{2-6}
    & \textbf{H = 32} & 1.43 & 4.00 & 4.91 & \textbf{1.40}\\
    \hline \hline
    \multirow{4}{*}{\vspace{-2.5em}\textbf{UCY}} &
    \textbf{\underline{H = 12}} & 0.65 & 2.49 & 2.57 & \textbf{0.49}\\
    \cline{2-6}
    & \textbf{H = 16} & 0.97 & 3.17 & 3.17 & \textbf{0.70}\\
    \cline{2-6}
    & \textbf{H = 20} & 1.22 & 4.20 & 4.18 & \textbf{0.95}\\
    \cline{2-6}
    & \textbf{H = 24} & 1.47 & 4.48 & 4.52 & \textbf{1.22}\\
    \cline{2-6}
    & \textbf{H = 28} & 1.72 & 4.60 & 4.68 & \textbf{1.50}\\
    \cline{2-6}
    & \textbf{H = 32} & 1.98 & 4.88 & 4.87 & \textbf{1.80}\\
    \hline
    \end{tabular}
  }
\end{table}

\begin{table}[t]
  \centering
  \caption{Mean Average Displacement (MAD) error when changing the observation period. Forecasting horizon is kept constant at 12 frames.}
  \label{table:table2Ab}
  \resizebox{\linewidth}{!}{
    \begin{tabular}{|c *{5}{|C{.075\textwidth}}|}
    \hline
    Dataset & Observation period & Vanilla LSTM & Social LSTM & MX-LSTM & Individual MX-LSTM \\
    \hline \hline
    \multirow{4}{*}{\vspace{-2.5em}\textbf{Zara 01}} &
    \textbf{O = 1} & 1.62 & {\bf 0.89} & 0.96 & 1.43 \\
    \cline{2-6}
    & \textbf{O = 4} & 0.90 & 0.69 & {\bf 0.64} & 0.79 \\
    \cline{2-6}
    & \textbf{\underline{O = 8}} & 0.90 & 0.68 & {\bf 0.59} & 0.72 \\
    \cline{2-6}
    & \textbf{O = 12} &0.90 &0.68 &{\bf 0.59} & 0.68 \\
    \cline{2-6}
    & \textbf{O = 16} & 0.90 & 0.68 & {\bf 0.59} & 0.60 \\
    \hline \hline
    \multirow{4}{*}{\vspace{-2.5em}\textbf{Zara 02}} &
    \textbf{O = 1} & 1.65 & 1.13 & {\bf 0.85} & 1.35 \\
    \cline{2-6}
    & \textbf{O = 4} & 1.17 & 0.74 & {\bf 0.48} & 0.84 \\
    \cline{2-6}
    & \textbf{\underline{O = 8}} & 1.09 & 0.63 & {\bf 0.35} & 0.63 \\
    \cline{2-6}
    & \textbf{O = 12} &1.01 &0.63 &{\bf 0.35} & 0.63 \\
    \cline{2-6}
    & \textbf{O = 16} & 0.99 & 0.63 & {\bf 0.33} & 0.62 \\
    \hline \hline
    \multirow{4}{*}{\vspace{-2.5em}\textbf{UCY}} &
    \textbf{O = 1} & 0.82 & 0.71 & {\bf 0.62} & 0.88 \\
    \cline{2-6}
    & \textbf{O = 4} & 0.65 & 0.63 & {\bf 0.49} & 0.59 \\
    \cline{2-6}
    & \textbf{\underline{O = 8}} & 0.67 & 0.62 & {\bf 0.49} & 0.53 \\
    \cline{2-6}
    & \textbf{O = 12} &0.65 &0.60 &{\bf 0.48} & 0.52 \\
    \cline{2-6}
    & \textbf{O = 16} & 0.63 & 0.60 & {\bf 0.48} & 0.52 \\
     \hline
    \end{tabular}
  }
\end{table}

\begin{table}[t]
  \centering
  \caption{Mean Average Displacement (MAD) error when changing the forecasting horizon. Observation interval is kept constant at 16 frames.}
  \label{table:table3Ab}
  \resizebox{\linewidth}{!}{
    \begin{tabular}{|c *{5}{|C{.075\textwidth}}|}
    \hline
    Dataset & Prediction interval & Vanilla LSTM & Social LSTM & MX-LSTM & Individual MX-LSTM \\
    \hline \hline
    \multirow{4}{*}{\vspace{-2.5em}\textbf{Zara 01}} &
    \textbf{Pred = 16} & 1.25 & 1.05 & {\bf 0.88} & 0.90 \\
    \cline{2-6}
    & \textbf{Pred = 20} & 1.27 & 1.46 & {\bf 1.19} & 1.26 \\
    \cline{2-6}
    & \textbf{Pred = 24} & 1.78 & 1.88 & {\bf 1.57} & 1.64 \\
    \cline{2-6}
    & \textbf{Pred = 28} & 2.39 &2.37 &{\bf 1.93} & 2.01 \\
    \cline{2-6}
    & \textbf{Pred = 32} & 3.09 & 3.00 & {\bf 2.32} & 2.57 \\
    \hline \hline
    \multirow{4}{*}{\vspace{-2.5em}\textbf{Zara 02}} &
    \textbf{Pred = 16} & 1.31 & 0.88 & {\bf 0.49} & 0.95 \\
    \cline{2-6}
    & \textbf{Pred = 20} & 1.87 & 1.24 & {\bf 0.67} & 1.28 \\
    \cline{2-6}
    & \textbf{Pred = 24} & 2.50 & 1.61 & {\bf 0.87} & 1.65 \\
    \cline{2-6}
    & \textbf{Pred = 28} &3.19 &2.05 &{\bf 1.11} & 2.04 \\
    \cline{2-6}
    & \textbf{Pred = 32} & 3.87 & 2.53 & {\bf 1.35} & 2.42 \\
    \hline \hline
    \multirow{4}{*}{\vspace{-2.5em}\textbf{UCY}} &
    \textbf{Pred = 16} & 1.02 & 0.80 & {\bf 0.71} & 0.72 \\
    \cline{2-6}
    & \textbf{Pred = 20} & 1.42 & 1.06 & {\bf 0.95} & 1.01 \\
    \cline{2-6}
    & \textbf{Pred = 24} & 1.87 & 1.34 & {\bf 1.2} & 1.40 \\
    \cline{2-6}
    & \textbf{Pred = 28} &2.37 &1.67 &{\bf 1.46} & 1.50 \\
    \cline{2-6}
    & \textbf{Pred = 32} & 2.92 & 2.21 & {\bf 1.80} & 1.90 \\
     \hline
    \end{tabular}
  }
\end{table}

We varied the observation interval, in order to understand how many frames are necessary to learn a meaningful representation of the trajectory, .
Table \ref{table:table2Ab} reports numerical results of an experiment where we kept the forecasting horizon fixed at 12 frames, and varied the observation period form 1 to 16 frames with the step size of 4 frames.
An observation period of 1 frame means we try to predict trajectories based only on a static observation of the individual, with no previous history taken into account. Results prove that one frame is not enough for all the methods under analysis. Despite this, the ranking of different approaches is maintained throughout all the experiments, with the only exception of Zara01 sequence with $O$=1, where Social LSTM outperforms competitors.
Interestingly, a rapid drop in error of about 30\% is obtained by observing 4 frames instead of 1. Furthermore, 8 frames are enough for the approaches to learn the overall shape of the trajectory in order to predict for the next 12 frames, as the error drop from observing 8 frames to 16 frames is below 1\%.

Finally, in order to understand in more depth how different methods perform for long term forecasting, we kept the observation interval constant at 16 frames and test increasing forecasting horizons. Table~\ref{table:table3Ab}, further validates the fact that 8 frames are sufficient for the LSTM approach to learn the representation of the trajectory. MX-LSTM is still the best performer but the error drop from observing 8 to observing 16 frames is negligible in long term forecatsing as well. This effect speaks about the capability of LSTM-based approaches. The performance already starts to saturate at 8 frames and adding more information does not bring the expected gain. In our view, this highlights the temporal modelling as one of the performance bottlenecks, on the way to progress in the field.


\subsection{Substitutes for Head Pose}

In this experiment, we analyze the importance of the head pose and question whether one may substitute it with more accessible proxies, such as the direction of the people pace. In more details,
we implement a Pace-MX-LSTM, which uses ground truth step directions instead of the head pose. Table~\ref{table:step} illustrates that having the step direction instead of the head pose downgrades the MX-LSTM, since positional data are already contained in the tracklet and the step direction can be extracted from the previous two positions. In fact, Pace-MX-LSTM gives consistently worse results.

In Table~\ref{table:step}, we additionally illustrate the importance of having access to manually annotated head poses during training. to study this aspect, we implemented the MX-LSTM-HPE-Train and Test,
where the head pose training data is given by a head-pose detector~\cite{hasan2017tiny}. As expected, MX-LSTM-HPE-Train and Test underperforms MX-LSTM and MX-LSTM-HPE (MX-LSTM-HPE is still trainned on manually labelled head poses, but it adopts a head pose estimator at inference time). This is especially so on Zara02, where conversational groups make the head pose estimation noisy due to the many partial occlusions. Still, MX-LSTM-HPE-Train
and Test remains comparable to prior state-of-the-art methods.

\begin{table}[t]
  \centering
  \caption{MAD errors on the different datasets}
  \label{table:step}
  \resizebox{\linewidth}{!}{
    \begin{tabular}{|l *{3}{|C{.11\textwidth}}|}
      \hline
      Dataset    & MX-LSTM  & MX-LSTM-HPE (Train and Test) & Pace-MX-LSTM \\ \hline
      \textbf{Zara01}     &  0.59   &  0.68       &    0.69      \\ \hline
      \textbf{Zara02}     &  0.35   &  0.51     &    0.73            \\ \hline
      \textbf{UCY}        &  0.49   &  0.58               &    0.59          \\ \hline
      \textbf{Town Centre} & 1.15   &  1.43               & 1.50                \\ \hline
    \end{tabular}
  }
\end{table}

\subsection{Qualitative Results}
\label{sec:qualitative}

\begin{figure*}[t!]
  \centering
  \includegraphics[width=1\linewidth]{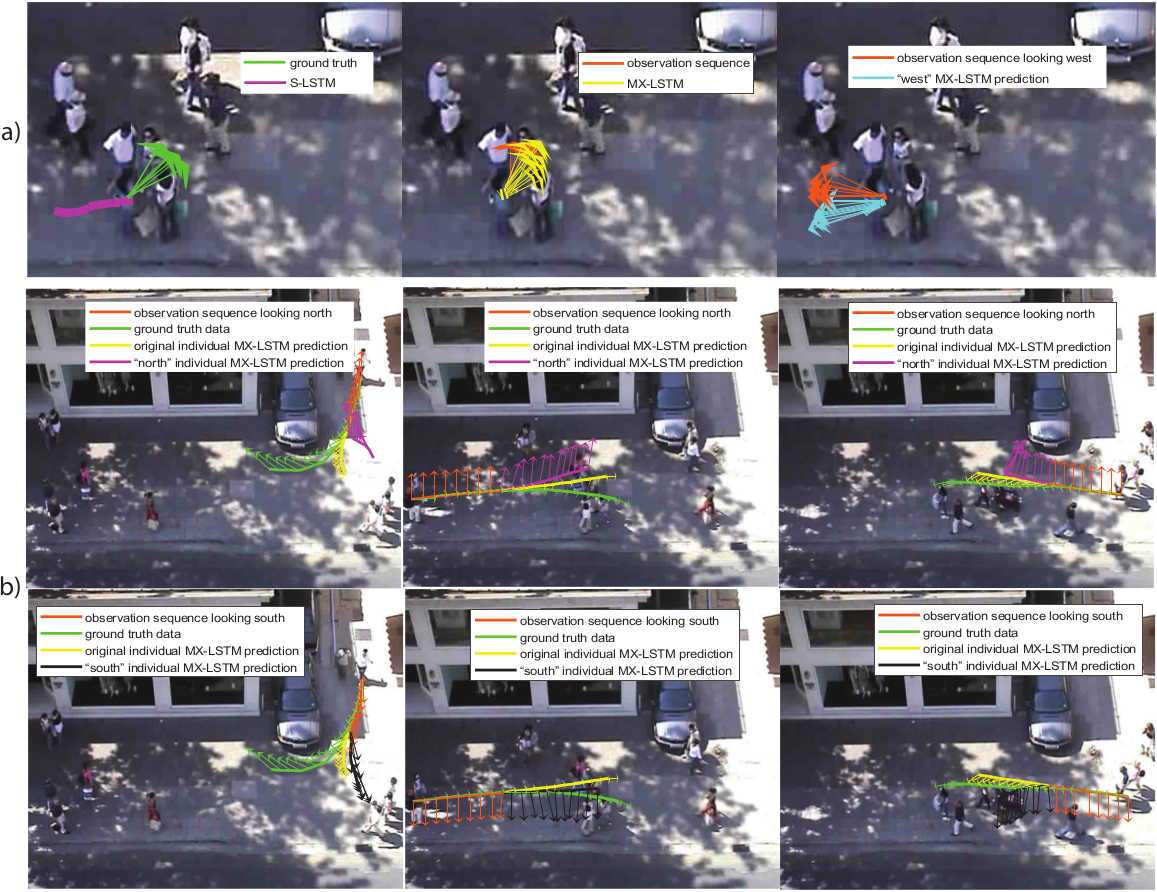}
  \caption{Qualitative results: a) MX-LSTM  b) Ablation qualitative study on Individual MX-LSTM (better in color).}
  \label{fig:Qualitative}
\end{figure*}

\figref{fig:Qualitative} shows  qualitative results on the Zara02 dataset, which was found as the most difficult throughout the quantitative experiments.
%
\figref{fig:Qualitative}a presents MX-LSTM results: a group scenario is taken into account, with the attention focused on the girl in the bottom-left corner. In the left column, the green ground-truth prediction vislets show that the girl is havign a conversation with the group members, nearly not moving at all, while moving her head around. The magenta curve (\figref{fig:Qualitative}a \textit{left}) represents the S-LSTM output, predicting erroneously that the girl would leave the group. This error confirms the problem of competing methods in forecasting the motion of people slowly moving or static, as discussed in Sec.~\ref{sec:motivation}. 
In the central column of \figref{fig:Qualitative}a, the observation sequence given to the MX-LSTM is shown in orange (almost static with oscillating vislets). The output prediction (yellow) shows oscillating vislets but no movement, confirming that the MX-LSTM has learnt this particular social behavior. If we provide the MX-LSTM with an artificial observation sequence with the annotated positions (real trajectory) but vislets oriented toward west (third column in \figref{fig:Qualitative}a, orange arrows), where no people are present, the MX-LSTM predicts a trajectory slowly departing from the group (cyan trajectory and arrows).


The two rows of \figref{fig:Qualitative}b analyze the Individual MX-LSTM, in which no social pooling is taken into account. Here pedestrians are not influenced by the surrounding people, and the forecast motion is only caused by the relationship between the tracklets and the vislets.
The first row in \figref{fig:Qualitative}b shows three situations in which the vislets of the observation sequence are manually altered to point north (orange arrows), thus orthogonal to the person trajectory. In this case the Individual MX-LSTM predicts a decelerating trajectory drifting toward north (magenta trajectory and vislets), especially visible in the second and third rows.  If the observation has the legit vislets (green arrows, barely visible since they are aligned with the trajectory), the resulting trajectory (yellow trajectory and vislets) has a different behavior, closer to the GT (green trajectory and vislets). Similarly, in the second row, we altered vislets to point to South. The prediction with the modified vislets is in black. The only difference is in the bottom left picture: here the observation vislets pointing south are in agreement with the movement, so that the resulting predicted trajectory is not decelerating as in the other cases, but accelerating toward south.


\section{Conclusion}
\label{sec:conc}

We have argued for the importance of people head poses, as encoded in the proposed \emph{vislets}, to forecast their future motion. We have shown that vislets are mostly aligned with the people motion, and therefore useful to forecast it. But when vislets are not aligned with the people motion, then they express the intention of people to change direction. Vislets differ from the current approaches, as most recent LSTM-based forecasting has only considered own and neighboring pedestrian positions. But this is close in spirit to decade-old works using the people desired goals. In this paper, the head pose is however estimated, not provided (e.g.\ by an oracle).

The use of vislets is enabled by the novel MX-LSTM framework. This jointly ``reasons'' on tracklets and vislets by means of a multi-variate Gaussian distribution, the covariance of which encodes the interplay of position and head pose. Our proposed log-cholesky parameterization allows its unconstrained optimization by the LSTM backpropagation, and it opens the way to including additional variables (\eg the people belonging to a social group).

Finally, this work has delved into a comprehensive evaluation of the proposed MX-LSTM, including ablation studies on vislets (both estimated and provided as GT), social pooling, view-frustum, observation and prediction time horizons. MX-LSTM provides currently state of the art performance and it is most effective when people slow down and look around to change direction, the Achilles heel of other current techniques.

\ifCLASSOPTIONcompsoc
  \section*{Acknowledgments}
\else
  \section*{Acknowledgment}
\fi
This project has received funding from the European Union's Horizon 2020 research and innovation programme under the Marie Sklodowska-Curie Grant Agreement No. 676455, and has been partially supported by the projects of the Italian Ministry of Education, Universities and Research (MIUR) "Dipartimenti di Eccellenza 2018-2022" and PORFESR 2014-2020 Work Program (Action 1.1.4, project No.10066183).


\ifCLASSOPTIONcaptionsoff
  \newpage
\fi



%


\bibliographystyle{ieee}
\bibliography{bare_adv}
%
\vspace{-26pt}
\begin{IEEEbiography}[{\includegraphics[width=1in,height=1.25in,clip,keepaspectratio]{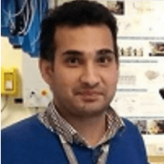}}]{Irtiza Hasan} is a research associate at Inception Institute of Artificial Intelligence (IIAI). His interests involve object detection and tracking, activity forecasting and pose estimation. He obtained his PhD from University of Verona, Italy in 2019. During his PhD, he was also a scientist working in Computer Vision Department at OSRAM (Munich, Germany). His PhD research focused on smart lighting and applying computer vision techniques to understand and forecast people activities.
\end{IEEEbiography}

\vspace{-20pt}

\begin{IEEEbiography}[{\includegraphics[width=1in,height=1.25in,clip,keepaspectratio]{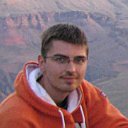}}]{Francesco Setti} is an Assistant Professor at the Department of Computer Science of the University of Verona working on the EU-H2020 project SARAS, and Associate Researcher of the Institute of Cognitive Science and Technology (ISTC-CNR). His research interests are in the application of machine learning and artificial intelligence techniques for industrial applications, with particular attention to the emerging fields of collaborative robotics and reinforcement learning for situation awareness decision making.
\end{IEEEbiography}

\vspace{-20pt}

\begin{IEEEbiography}[{\includegraphics[width=1in,height=1.25in,clip,keepaspectratio]{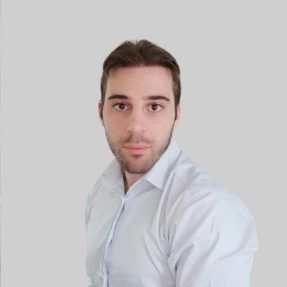}}]{Theodore Tsesmelis} is currently a computer vision researcher at the Computer Vision Department at OSRAM GmbH in Germany. Currently he is applying research and development regarding smart workspace management with the use of computer vision and machine learning. Other research interests involve reconstruction and recognition of 3D scenes, the scene material and object properties, and the estimation of lighting propagation.
\end{IEEEbiography}

\vspace{-20pt}

\begin{IEEEbiography}[{\includegraphics[width=1in,height=1.25in,clip,keepaspectratio]{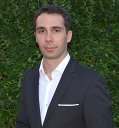}}]{Vasileios Belagiannis} is Junior-Professor at University of Ulm. He holds a degree in engineering from Democritus University of Thrace and M.Sc. in Computational Science and Engineering from TU M{\"u}nchen. He completed his doctoral studies at TU M{\"u}nchen and continued as post-doctoral research assistant at University of Oxford (Visual Geometry Group). Prior to joining University of Ulm, he was conducting research at OSRAM in Germany. His research is focused on deep learning, machine learning and computer vision, including applications from autonomous driving.
\end{IEEEbiography}

\vspace{-20pt}

\begin{IEEEbiography}[{\includegraphics[width=1in,height=1.25in,clip,keepaspectratio]{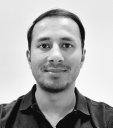}}]{Sikandar Amin} is a senior computer vision scientist at OSRAM in Germany, where he manages multiple AI projects regarding infrastructure sensing platforms for autonomuos driving, and indoor smart lighting applications.
His R\&D interests include object detection, tracking, re-identification and full body pose estimation. He obtained his PhD from TU M{\"u}nchen in computer vision. His PhD research includes 2D and 3D human pose estimation for higher level tasks including activity recognition and studying human emotions during dyadic interactions in complex real-world settings.

\end{IEEEbiography}

\vspace{-20pt}

\begin{IEEEbiography}[{\includegraphics[width=1in,height=1.25in,clip,keepaspectratio]{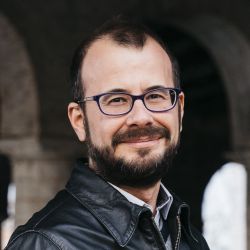}}]{Alessio Del Bue} is a tenured senior researcher at the PAVIS (Pattern Analyisis and computer VISion) Department of the Italian Institute of Technology (IIT) where he is leading the Visual Geometry and Modelling (VGM) Lab. His current research interests are related to 3D scene understanding from images and sound, 3D digitisation technology for Cultural Heritage studies, non-rigid image registration and reconstruction, and sensors/targets localisation and room reconstruction from sound.
\end{IEEEbiography}

\vspace{-20pt}

\begin{IEEEbiography}[{\includegraphics[width=1in,height=1.25in,clip,keepaspectratio]{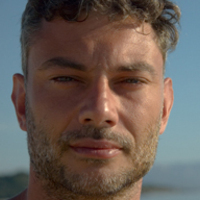}}]{Marco Cristani} is an Associate Professor (Professore Associato) at Computer Science Department, University of Verona. His main research interests are in statistical pattern recognition (mainly deep learning and generative modeling) and computer vision, with emphasis on social signal processing, \ie, how to model human activities with computer vision tools, following social psychology principia. In particular, he is interested in fashion modeling (clothing parsing, recommendation, attribute learning) and how fashion is related to personality. Other interests are on video surveillance applications, such as human activity modeling, people re-identification and pedestrian detection.
\end{IEEEbiography}

\vspace{-20pt}

\begin{IEEEbiography}[{\includegraphics[width=1in,height=1.25in,clip,keepaspectratio]{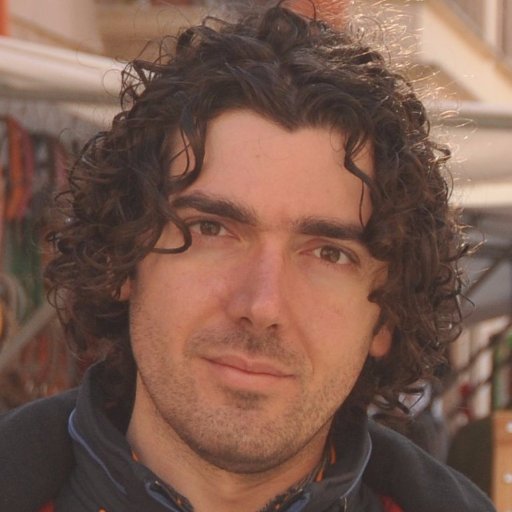}}]{Fabio Galasso} heads the Computer Vision Department at OSRAM (Munich, Germany), an international team conducting R\&D in artificial intelligence, computer vision and machine learning, in relation to smart lighting applications. Prior to OSRAM, he has conducted research at the University of Cambridge (UK) and at the Max Planck Institute for Informatics (Germany). He received his Master's Degree cum laude from the RomaTre University (Italy) and his PhD from the University of Cambridge (UK).
\end{IEEEbiography}

\vfill








\end{document}